\documentclass[conference]{IEEEtran}
\IEEEoverridecommandlockouts

\usepackage{cite}
\usepackage{amsmath,amssymb,amsfonts}
\usepackage{graphicx}
\usepackage{url}
\usepackage{times}
\usepackage[labelformat=simple]{subcaption}

\usepackage{pstricks-add}
\usepackage[inline]{enumitem}
\usepackage{booktabs,colortbl, array}
\usepackage[OT1]{fontenc} 

\begin{document}

\title{Modeling Pharmacological Effects with\\Multi-Relation Unsupervised Graph Embedding\\
\thanks{We thank the partial support given by the Project: Models, Algorithms and Systems for the Web (grant FAPEMIG / PRONEX / MASWeb APQ-01400-14), and authors' individual grants and scholarships from CNPq and Kunumi.}
}

\author{\IEEEauthorblockN{Dehua Chen}
\IEEEauthorblockA{
\textit{CS Dept. UFMG \& Kunumi}\\
Belo Horizonte, Brazil\\
chendehua@dcc.ufmg.br}
\and
\IEEEauthorblockN{Amir Jalilifard}
\IEEEauthorblockA{
\textit{CS Dept. UFMG}\\
Belo Horizonte, Brazil\\
jalilifard@ufmg.br}
\and
\IEEEauthorblockN{Adriano Veloso}
\IEEEauthorblockA{
\textit{CS Dept. UFMG}\\
Belo Horizonte, Brazil\\
adrianov@dcc.ufmg.br}
\and
\IEEEauthorblockN{Nivio Ziviani}
\IEEEauthorblockA{
\textit{CS Dept. UFMG \& Kunumi}\\
Belo Horizonte, Brazil\\
nivio@dcc.ufmg.br}
}

\maketitle

\begin{abstract}
A pharmacological effect of a drug on cells, organs and systems refers to the specific biochemical interaction produced by a drug substance, which is called its mechanism of action. Drug repositioning (or drug repurposing) is a fundamental problem for the identification of new opportunities for the use of already approved or failed drugs. In this paper, we present a method based on a multi-relation unsupervised graph embedding model that learns latent representations for drugs and diseases so that the distance between these representations reveals repositioning opportunities. Once representations for drugs and diseases are obtained we learn the likelihood of new links (that is, new indications) between drugs and diseases. Known drug indications are used for learning a model that predicts potential indications. Compared with existing unsupervised graph embedding methods our method shows superior prediction performance in terms of area under the ROC curve, and we present examples of repositioning opportunities found on recent biomedical literature that were also predicted by our method.
\end{abstract}


\graphicspath{{img/}, {pstricks/}}

\section{Introduction}
\label{intro}

Drug repositioning (aka repurposing) can be defined as renewing failed drugs and expanding successful ones by developing new therapeutic uses that are beyond their original uses or initial approved indications. Repositioned drugs account for approximately 30\% of the US Food and Drug Administration (FDA) approved drugs in recent years~\cite{fda}. A repositioned drug uses de-risked compounds, going directly to preclinical testing and clinical trials, thus providing inexpensive alternatives to the costly pipeline associated with the development of new drugs. One of the well-known examples is sildenafil citrate (brand name: Viagra), which was repositioned from a common hypertension drug to a therapy for erectile dysfunction~\cite{viagra}.

\begin{figure}
\centering
\resizebox{\linewidth}{!}{
    \includegraphics{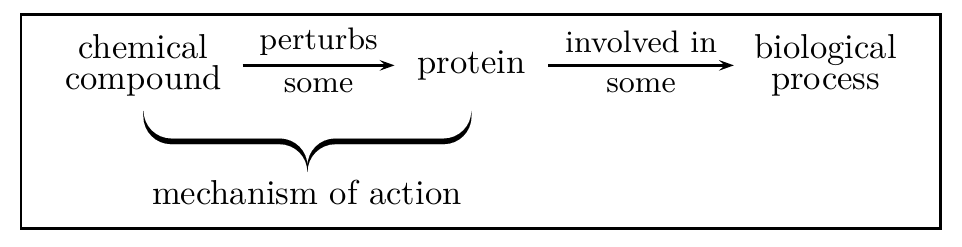}
}
\caption{A chemical is assigned as a treatment to a disease because it exhibits a particular mechanism of action that affects biological processes associated with the disease. Repositioning opportunities exist because the same drug perturbs multiple proteins themselves involved in multiple biological processes.}
\label{fig:moa}
\end{figure}

Figure~\ref{fig:moa} illustrates the biochemical interaction that gives rise to the pharmacological effect of a drug. This paper is motivated by the problem of finding drug repositioning opportunities by modeling the mechanisms of action of drugs~\cite{moa}. For instance, different biological solutions might be considered in order to chemically decrease the blood pressure such as removing the excess of salt from the body, thereby decreasing the tension in the vessels, or inhibiting the vasoconstrictive signalling of a hormone, or acting directly on the cells physically narrowing the vessels and preventing their unwanted action this way~\cite{ong}. Each of the aforementioned solutions requires a different mechanism of action. The same drug can have several mechanisms of action and therefore it can potentially play a multitude of roles by perturbing proteins involved in various biological processes, which are accountable for the drug polypharmacology~\cite{polypharmacy}. Thus, drug repositioning is a direct application of drug polypharmacology~\cite{poly1}.

\begin{figure}[htb]
\centering
\resizebox{\linewidth}{!}{
    \includegraphics{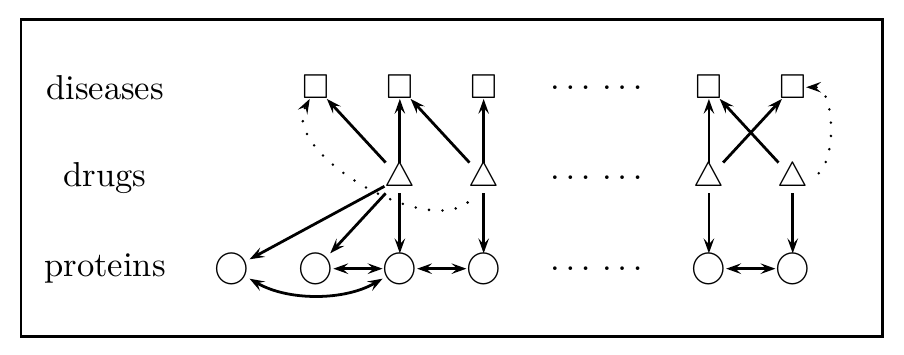}
}
\caption{Multi-relation graph, composed of drug-protein, drug-disease and protein-protein interactions. A drug (\small{$\triangle$}) perturbs some proteins (\small{$\bigcirc$}) and this drug is indicated to certain diseases (\small{$\square$}). A single drug may perturb different proteins, and these proteins may also interact. Further, the same drug may be indicated to different diseases. Links may provide evidence for repositioning opportunities (i.e., dotted links).}
\label{fig:graph}
\end{figure}

Our main goal is to discover new relations between current drugs and diseases by utilising existing public drug-disease-protein interactions. The main three steps of our proposed method are as follows. In the first step, we built a large and heterogeneous graph comprising drug, disease, and protein entities that are linked according to information collected from the biomedical literature, as shown in Figure~\ref{fig:graph}. Specifically, we formulate the drug repositioning problem as a three layer multi-relation directed graph $\mathcal{G} = (\mathcal{V}, \mathcal{R}, E)$, where $\mathcal{V}$ is the set of entities (i.e., drugs, diseases and proteins), $\mathcal{R}$ is a set of relations (i.e., drug-protein, drug-disease and protein-protein), and $E$ is a set of edges connecting different entities in $\mathcal{V}$. In the graph, mechanisms of action are represented by relations involving drugs and proteins and repositioning opportunities are represented by (hidden) relations involving drugs and diseases. The graph also contains protein-protein interactions in order to increase connectivity and information propagation while learning node representations. The datasets used to build the graph are described in Section~\ref{sec:data}.

In the second step, our goal is to find a low-dimensional latent representation for drugs and diseases, so that the latent representation embeds the relationship between mechanisms of action and drug indications. Drug-protein and drug-disease interaction graphs usually exhibit a particular structure with many isolated sub-graphs, and often a protein is linked to drugs residing in different parts of a graph. We employ a SkipGram based algorithm~\cite{skipgram} to learn node representations in an unsupervised way, but instead of performing deep random walks to produce contexts~\cite{node2vec,deepwalk}, we employ a restricted number of permutations over the immediate neighborhood of a node as context to generate its representation~\cite{nbne}. This choice is motivated by the particularly sparse structure of drug-disease and drug-protein interaction graphs. Further, we exploit the multi-relation nature of the graph by employing two types of contexts while learning node representations: contexts  composed of drugs and proteins (i.e., mechanisms of action) and contexts composed of drugs and diseases (i.e., drug indications). This results in an embedding for each drug and for each disease, so that adjacent entities are placed close to each other in the vector space, while unconnected entities are pushed apart. As a result, drugs and diseases that have a similar distribution of neighbors will end up being nearby in the vector space. Details for obtaining node embeddings can be seen in Section~\ref{sec:node_embedding}.

In the third step, we learn the likelihood of new links between drugs and diseases, as representations for drugs and diseases were obtained in previous step. Known drug indications are used for learning a parametric model which predicts other likely indications. Our evaluation follows the typical cross-validation framework, in which a subset of the known drug uses are hidden. Details of the experimental setup can be seen in Section~\ref{experimental_setup} and the results obtained by different embedding algorithms we used can be seen in Section~\ref{results}.

In the following we briefly summarize our contributions:
\begin{itemize}
\item We employ interaction graphs involving drugs, diseases and proteins in order to learn suitable vector representations for drugs and diseases. The input graph presents particular characteristics, such as high sparsity and low connectivity, so that contextual information based on the immediate neighborhood is likely to produce better representations than typical random walk approaches.
\item Given the vector representations for drugs and diseases, we build a parametric model learned to identify if a specific drug is indicated to a specific disease.
\item We evaluate the effectiveness of our model on predicting repositioning opportunities under a cross-validation framework. Our model reaches an area under the curve of $+$0.98, being significantly superior than predictive models built using contextual information produced by deep random walks.
\item Finally, we compared our specific findings with reposition opportunities reported in the recent biomedical literature. We present some interesting cases that were predicted by our model, including the use of Amitriptyline for relief of Fibromyalgia in adults. Amitriptyline is an antidepressant that is now being reported in the literature to treat Fibromyalgia if used at doses below those at which the drugs act as antidepressants.
\end{itemize}

\section{Related Work} \label{sec:related_work}

Several strategies have been proposed for drug repositioning: (i) structure-based, (ii) repositioning based on transcriptional signatures, (iii) ligand-based, and (iv) network-based. Structured-based methods follow the idea that similar proteins have similar functionality. Accordingly, this similarity comparison can be used to find secondary targets of already existing drugs \cite{ehrt2016impact}. Molecular transcriptional signatures can be compared to create relations between drugs and new disease indications. These relations provide useful information for finding new uses of known drugs \cite{lamb2006connectivity}. Ligand-based approaches are based upon the concept that similar compounds tend to have similar biological properties. In drug repositioning, this method has been widely used to analyze and identify the activity of ligands for new disease indications \cite{liu2010pharmmapper}.

In recent years, machine learning has been vastly employed in drug repositioning. Authors in \cite{kumar2019drug} used a fully connected deep neural network for training the model using transcriptional data at gene level to predict drug therapeutics and to use them in drug repositioning. They analyzed the confusion matrix and found out that the misclassified cases can indeed be considered as an indication of their potential in novel uses. Donner et al. \cite{donner2018drug} proposed ligand-based approach based on the learning of embeddings of gene expression profiles using deep neural networks and considered it as a measure of compound functional similarity for drug repositioning. Hu and Agarwal \cite{hu2009human} created a drug-disease network using publicly available gene expression.

Graphs are the typical structures used to model the relationships between drugs and diseases. The major challenge is to find a way to incorporate complex structures like graphs into the existing machine learning algorithms. Thereby, several models were proposed for learning a low dimensional representation of graphs known as graph embedding \cite{angles2008survey,belkin2002laplacian,node2vec,kipf2016semi}. In order to model the polypharmacy side-effect, Zitnik et al. \cite{zitnik2018modeling} trained graph convolutional neural networks, with proteins and drugs as nodes and drug-protein and drug-drug interaction as edges. Deepika and Geetha \cite{deepika2018meta} used node2vec \cite{node2vec} representations along with bagging Support Vector Machine (SVM) to predict drug-drug interactions. Gao et al. \cite{gao2018interpretable} applied Long Short-term Memory Neural Networks (LSTMs) and graph-based convolutional neural network to obtain a low dimensional representation of protein and drug structures. These representations were then engaged in the prediction of drug-target interactions.
Yamanishi et al. \cite{yamanishi} introduced a bipartite graph-learning method based on kernel regression in order to learn a co-mapping of drugs and proteins into a common pharmacological space. In the pharmacological space, the correlation between compound-protein pairs can be conveniently calculated to predict their interactions for drug repositioning. \cite{zheng} proposed a method to factorize the existing drug-target relations so as to predict the new relations constrained by the drug-drug and disease-disease similarity networks. Finally, \cite{xia} proposed a semi-supervised learning method in which two classifiers in drug and disease space are learned and then combined together to give a final score for drug-disease interaction prediction.

Recent representation learning methods include neural fingerprints~\cite{fingerprint}, graph convolutional networks~\cite{gcn}, and message passing networks~\cite{chemistry}. However, these graph embedding methods do not apply in our setting, since they solve a supervised graph classification task and/or embed entire graphs while we embed individual nodes.

\section{Data} \label{sec:data}

\begin{table}[t]
\centering
 \caption{Basic statistics of the data.}
 \begin{tabular}{lr}
    \multicolumn{2}{c}{drug-protein}\\
    \hline\\
    \# of drugs & 584 \\
    \# of proteins & 16,546 \\
    \# of interactions & 1,824,204
 \end{tabular}
 \bigskip
 \begin{tabular}{lr}
    drug-disease & known indications\\
    \hline\\
    \# of drugs & 600 \\
    \# of diseases & 508 \\
    \# of interactions & 2,836 
 \end{tabular}
 \label{tab:stats}
\end{table}

In this section, we discuss the datasets used to build the graph presented in Figure~\ref{fig:graph}.
As in~\cite{zitnik2018modeling}, we used the human protein-protein interaction (PPI) network compiled by~\cite{menche,biogrid}, integrated with additional PPI information from~\cite{damian}. The PPI graph contains physical interactions experimentally documented in humans, such as metabolic enzyme-coupled interactions and signaling interactions. The network is unweighted and undirected with 19,085 proteins and 719,402 physical interactions. Table \ref{tab:stats} presents statistics about the data from which we built two graphs:
\begin{enumerate}
  \item 
For the graph drug-protein, we obtained relationships between drugs and proteins from the STITCH database~\cite{biogrid}. This database integrates various chemical and protein networks and there were over 8,083,600 interactions present between 8,934 proteins and 519,022 chemicals. We considered only the interactions between chemicals (i.e., drugs) and proteins that had been experimentally verified, which comprises 16,546 proteins and 584 drugs, and there are 1,824,204 interactions amongst them.
 \item 
Drugbank~\cite{drugbank} was used to retrieve known drug-disease links. DrugBank is a bioinformatics and cheminformatics resource that provides a knowledge-base for drugs, drug actions and drug targets. We focused on 600 drugs that were indicated to 508 diseases, resulting in a total of 2,836 drug-disease links.  

\end{enumerate}

\section{Unsupervised Node Embedding}
\label{sec:node_embedding}

In this section we  aim to learn representations for drugs and diseases that best preserve the original graph structure, generalizing mechanisms of action in order to find novel uses and repositioning opportunities. Graph embedding consists in finding a continuous vector space representation for entities in the set of nodes $\mathcal{V}$. The task is to learn a dictionary $Z\in\mathbb{R}^{|\mathcal{V}|\times d}$, with one $d-$dimensional embedding for each node in $\mathcal{V}$. In other words, graph embeddings are the transformation of a graph to a set of vectors, by capturing the graph structure as well as node-to-node relationship. Unsupervised learning of graph embeddings has benefited from the information contained in contexts~\cite{nbne}, and thus embedding methods usually work by simulating contexts and operate in two steps:

\begin{enumerate}
\item They sample pair-wise relationships from the graph through random walks. Each random walk generates a sequence of nodes, simulating a context.
\item They train an embedding model, e.g. using Skipgram algorithm~\cite{skipgram}, to learn representations that encode pairwise node similarities.
\end{enumerate}

Embedding methods differentiate mainly on the first step, as there are many possible ways to extract context from a graph. The best strategy for producing context depends on specific characteristics of the graph. In this work the contexts are based solely on the first order neighborhoods of nodes, defined here as the nodes that are directly connected. Consequently, nodes' representations will be mainly defined by their first order neighborhoods and nodes with similar neighborhoods (contexts) will be associated with similar representations. This results in embeddings focused mainly on the first-order proximity. More specifically, we first separate a node neighborhood in small groups and then we maximize the log likelihood of predicting a node given another in such a group~\cite{ref5}.

\subsection{Generating Contextual Groups} \label{subsection_sentence}

The first step is to group nodes based on their neighborhoods, so that context can be exploited. There are two main challenges in forming groups from neighborhoods, as follows:

\begin{itemize}
\item Nodes have different degrees, so groups containing all the neighbors from a node are difficult to treat.
\item There is no explicit order in the nodes in a neighborhood. So there is no clear way to choose the order in which they would appear in a group.
\end{itemize}

To deal with these challenges, we create small groups with only $k$ neighbors in each, using random permutations of their neighborhoods~\cite{nbne}.
The number of permutations $n$ is specified and controls the trade-off between training time and increasing the training dataset. Selecting a higher value for $n$ creates a more uniform distribution on possible neighborhood groups, but also increases training time.

\subsection{Learning Representations} \label{subsec:learning_representations}

The first step results in a set of groups $S$, where each member of $S$ is a subset of nodes in the graph. Then, we learn vector representations of nodes by maximizing the log likelihood of predicting a node given another node in a group and given a set of representations $r$, making each node in a group predict all the others. The log likelihood to maximize is given by:
\begin{equation} \label{sg_entropy}
  \max\limits_{r} \quad \frac{1}{|S|}  \sum_{s \in S} \left( \log \left( p\left( s | r \right) \right)\right)
\end{equation}
\noindent where $p\left( s | r \right)$ is the probability of each group, given as:
\begin{equation} \label{eq:sentence_prob}
  \log \left( p\left( s | r \right)\right) = \frac{1}{|s|}  \sum_{v_i \in s} \left( \sum_{v_j \in s, v_j \neq v_i} \left( \log \left( p\left( v_{j} | v_i, r \right) \right) \right)\right)
\end{equation}
\noindent where $v_i$ is a node in the graph and $v_{j}$ are the other nodes in the same group. The probabilities in this model are learned using the feature vectors $r_{v_i}$, which are then used as the node representations. The probability $p\left( v_{j} | v_i, r \right)$ is given by:
\begin{equation} \label{eq:prob_node}
  p\left( v_j | v_i, r \right) = \frac{\exp \left( r_{v_j}^{T} \times r_{v_i} \right) }{ \sum_{v \in V} \left( \exp \left( r_{v}^{T} \times r_{v_i} \right) \right) }
\end{equation}
where $r^T_{v_j}$ is the transposed output feature vector of node $j$, used to make predictions. The representations $r_{v}$ and $r_{v}$ are learned simultaneously by optimizing Equation~\ref{sg_entropy}.
Essentially, by optimizing this log probability the algorithm maximizes the likelihood of predicting a neighbor given a node, creating node embeddings so that nodes with similar neighborhoods have similar representations~\cite{ref8}. Since there is more than one neighbor in each group, this model also makes connected nodes having similar representations, because they will both predict each others neighbors, resulting in representations also with first order similarities. A trade-off between first and second order proximities can be achieved by changing the parameter $k$, which controls the number of nodes within each group.

\section{Experimental Setup}
\label{experimental_setup}

Our data is a multi-relation graph compose of drug-protein and drug-disease interactions. Thus, in order to learn our models, it is necessary to select an efficient embedding space in order to better exploit the information within the graph. We discuss the choice for an appropriate embedding space which includes evaluating different graph-embedding algorithms and their corresponding hyper-parameters.

\subsection{Learning the Embedding Space}

We first find an efficient embedding space for the different node embedding algorithms that will be compared in our experiments. This involves 25 hyperparameter combinations that were randomly selected for each algorithm and embedding models are then learned in an unsupervised way. We considered three graph-embedding algorithms in our experiments: (i) DeepWalk~\cite{deepwalk}, where the best hyperparameters are: window size of 12, number and length of walks were set to 7 and 25, respectively; (ii) Node2Vec~\cite{node2vec}, with window size, number and length of walks equal to 5, 57 and 73, respectively; (iii) NBNE~\cite{nbne}, with hyperparameters: window size of 6 and number of permutations set to 30.

\subsection{Model Evaluation}

We used the Multilayer Perceptron (MLP) as a binary classifier, which predicts possible links between drugs and diseases using the embedding space. Specifically, the vector of a drug and the vector of a possible indication (i.e., a disease) to the drug are concatenated, and the MLP model takes the final vector as input and makes a prediction (in this case, the output is the probability of a link existing between the drug and the disease). As shown in Table~\ref{tab:stats}, the known indications that form the drug-disease interaction graph contains 2,836 links. We used 5-fold cross-validation to assess the embedding's quality. Thus, we divided 2,836 into five folds, each time one of them is used for the validation and the rest for the training. As the known indications data contains only positive examples, we have generated 30,196 negative examples using the complementary graph of the known indications in order to learn the MLP model~\cite{ref3}, as shown in Table~\ref{tab:neg}. It is worth mentioning that there are more negative occurrences than positive ones in the real-world scenario, because drugs are produced for a small group of diseases or health issues, being ineffective for others. As can be seen in the table, the data is highly imbalanced, thus making our experiment close to a real-world scenario. Finaly, we used \textit{area under the curve} (AUROC score) as the basic measure for assessing the performance of the algorithms~\cite{ref6}.

\begin{table}[t]
\centering
\caption{Number of positive and negative examples used to learn the embedding space and to train the parametric model.}
\resizebox{\linewidth}{!}{
 \begin{tabular}{l|r|rr}
    Steps & interactions & {examples} \\
    \hline
    \\
    Learning Embedding Space & 1,827,040 & $-$ \\[2pt]
    Model Evaluation & $-$ & 2,836 ($+$) 30,196 (-)
 \end{tabular}
}
 \label{tab:neg}
\end{table}

\section{Results}
\label{results}

In this section we report results obtained by the three embedding algorithms DeepWalk, Node2Vec and NBNE. We also discuss examples of drug repositioning opportunities endorsed by recent biomedical literature.

\begin{figure}[ht]
    \centering
    \resizebox{0.5\textwidth}{!}{\input{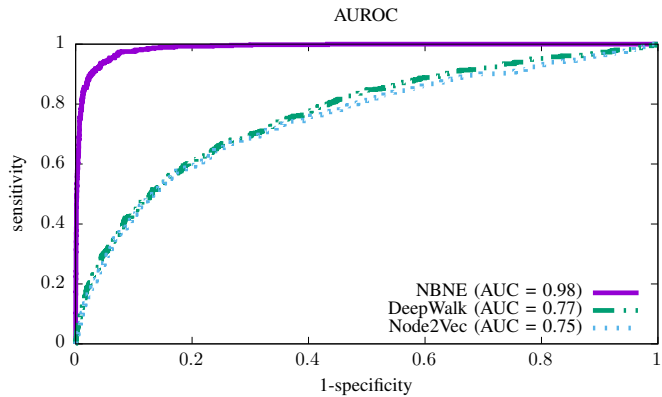}}
    \caption{AUROC values for Deepwalk, Node2vec and NBNE.}
    \label{fig:AUC}
\end{figure}

\vspace{0.1in}
\noindent\textbf{Prediction Performance: }
As shown in Figure \ref{fig:AUC}, NBNE has obtained the best result. Specifically, NBNE achieved numbers as high as 0.98 in terms of AUROC, while Node2vec and Deepwalk achieved 0.75 and 0.77, respectively. The improvement provided by NBNE compared to Deepwalk and Node2Vec is significant $-$ 27\% and 28\% of improvement respectively. The main difference of NBNE from the other two algorithms is the context generation approach, as NBNE is based on the neighborhood while the other two algorithms are on random walks, as discussed in Section~\ref{subsection_sentence}. It seems that the neighborhood based approach generates more accurate representation in the drug repositioning scenario.

\begin{figure*}[h!]
    \centering
    \resizebox{0.65\textwidth}{!}{\input{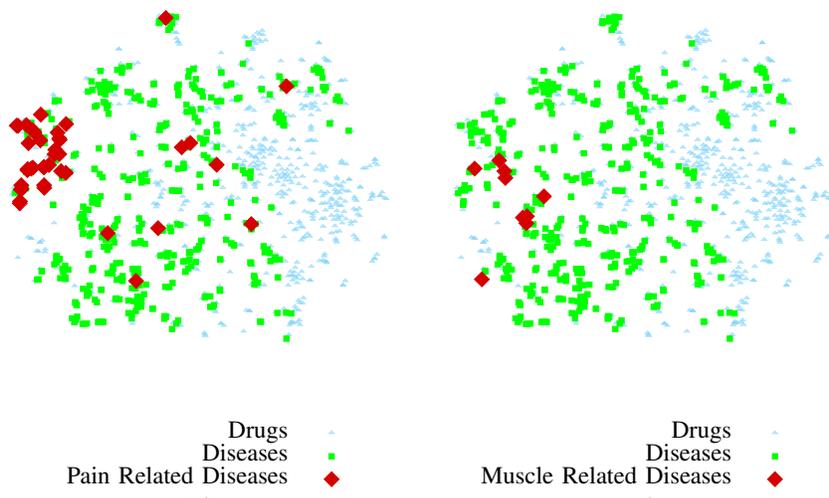}}
    \caption{Proximity of related diseases. Left $-$ Pain related diseases. Right $-$ Muscle related diseases.}
    \label{fig:related}
\end{figure*}

\begin{figure*}[h]
    \centering
    \resizebox{0.65\textwidth}{!}{\input{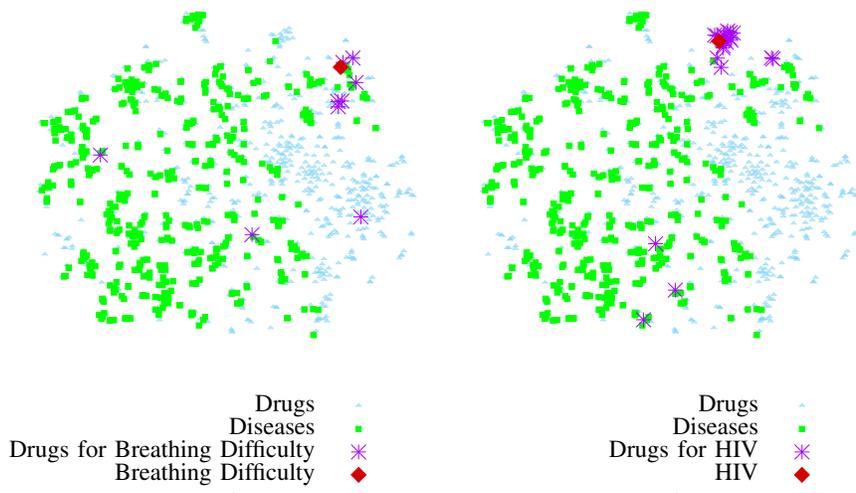}}
    \caption{Proximity of diseases and their corresponding medications. Left $-$ \textit{Breathing Difficulty}. Right $-$ \textit{HIV}. In general, drug indications and the corresponding health problems are located closely.}
    \label{fig:disease_med1}
\end{figure*}

\begin{figure}[h!]
\vspace{-0.3in}
    \centering
    \label{fig:clusters}
\resizebox{\linewidth}{!}{
    \includegraphics{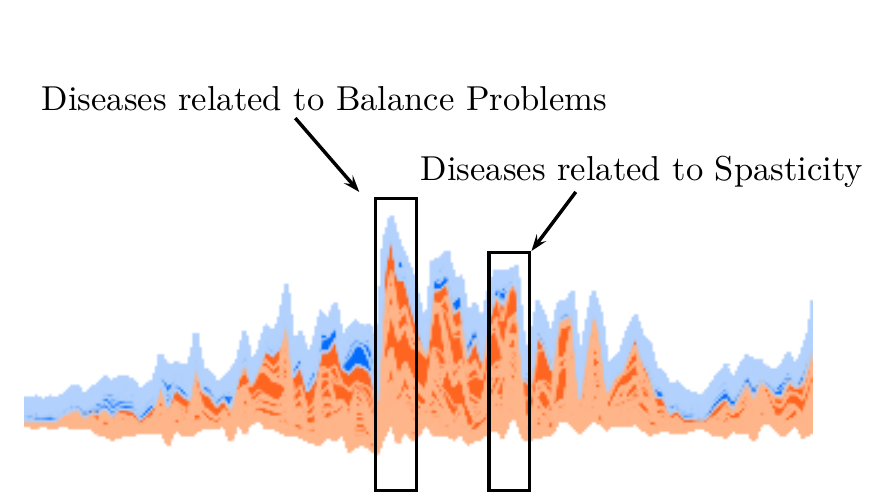}
}
\caption{The x$-$axis represents the instances and the y$-$axis is the corresponding SHAP explanation. Predictions that have similar explanation are placed next to each other in the x$-$axis.}
\label{fig:shap}
\end{figure}

\vspace{0.1in}
\noindent\textbf{Related Diseases in the Embedding Space:}
We employ t-SNE in order to visualize the embedding space of drugs and diseases. T-SNE is a technique used to visualize high-dimensional data by giving each data point a location in a two-dimensional map~\cite{tsne}. The visualization suggests some insights about the reasons that lead to the good performance of our method. In order to have a clear visualization, the drugs are represented by triangles (blue) and the diseases by rectangles (green). Further, some points are also highlighted in the figures to demonstrate interesting properties of the embedding space. Figure \ref{fig:related} shows that similar diseases have close vector representations. Figure \ref{fig:related} (Left) shows a cluster of diseases (red points) representing pain related diseases, while Figure~\ref{fig:related} (Right) shows a cluster of muscle related diseases. These visualizations suggest that our method generates meaningful representations as related diseases are located close to each other in the embedding space.  

\vspace{0.1in}
\noindent\textbf{Diseases and their Corresponding Drugs in the Embedding Space: }
We have also analysed the spatial relation of diseases and their corresponding drugs. Figure \ref{fig:disease_med1} (Left) highlights drugs which are used to treat \textit{Breathing Difficulty}. In this case, most of the indicated drugs are concentrated next to the disease. The same trend is observed in Figure~\ref{fig:disease_med1} (Right), where we highlighted the disease \textit{HIV} $-$ again, most of the indicated drugs are placed next to the disease. These visualization give us a great insight of the embedding space generated by NBNE.

\vspace{0.1in}
\noindent\textbf{Model Explainability: } Features within our embeddings are not meaningful as we do not know what each feature is actually representing. Fortunatelly, in order to better understand model predictions, we can still group the predictions based on the features used by the model while performing the prediction~\cite{ref10}. Intuitively, positive predictions involving the same disease should be located next to each other, as the corresponding drugs may share the same mechanism-of-action. We employ SHAP (SHapley Additive exPlanations~\cite{shap}) in order to calculate the importance of each feature for each prediction. Figure~\ref{fig:shap} shows some predictions performed by our model, so that predictions associated with similar feature importances are placed next to each other along the x$-$axis. For each prediction, a feature is associated either with the blue or the red color. Features associated with the blue color are contributing to decrease the probability of the disease being a new use for the drug. Therefore, a positive prediction is always observed when the red color is the dominant one. We highlight two cases in the figure, showing positive predictions for the same diseases. Specifically, we found a cluster involving different drugs being predicted for \textit{Balance Problems}, and another involving different drugs being prodicted for \textit{Spasticity}.

We found many other clusters, some of them composed by different diseases. One of these clusters shows that Bipolar II disorder and headaches respond equally to some specific embedding features. Interestingly, a recent research confirms that migraine with active headache is associated with other painful physical symptoms among patients with major depressive disorder \cite{hung2019migraine}. In another example, Multiple Sclerosis (MS) and Toxoplasmosis were placed into the same cluster. According to a study carried put by  \cite{enriquez2017cerebral}, Toxoplasmosis should be considered as differential diagnosis of tumefactive MS. Another interesting result is the association between panic disorder and Amyotrophic Lateral Sclerosis (ALS) found by our prediction model and recently confirmed by \cite{siciliano2019assessing}. According to this study, 33\% of patients who suffer from ALS also suffer from some kind of panic and anxiety disorder.

\vspace{0.1in}
\noindent\textbf{Repositioning Opportunities and Biomedical Literature: } Table \ref{table:table3} presents examples of repositioning opportunities. Our prediction model suggests Gabapentin as a candidate for bipolar II disorder, which has been confirmed by \cite{fullerton2010rise} and in several other studies. While Naproxen is used in treating balance problems, it can also be used for treating Myofascial Pain~\cite{khalighi2016low}, which is confirmed by our model as it places these diseases and medications in the same group. Recent studies show that fibromyalgia is associated with muscle tension and depression \cite{bosco2019clinical}. Recent research carried out by \cite{guymer2019pharmacological} shows that Amitriptyline, which has been used in the treatment of muscle tension, is a possible candidate for fibromyalgia. Lately, \cite{bubnova2019effects} confirmed that both Amlodipine and Atorvastatin caused significant improvement in patients with high blood pressure which is in accordance with our results.

\begin{table}[t]
\centering
\caption{List of some possible candidatesfor drug repositioning reported in biomedical literature and found by our algorithm.}
\begin{tabular}{rrr}
\toprule 
Medication & Target disease & Also appeared in\\
\midrule
\rowcolor{black!20} Gabapentin  & Bipolar II disorder & \cite{fullerton2010rise} \\
Naproxen         & Myofascial Pain     & \cite{khalighi2016low} \\
\rowcolor{black!20} Amitriptyline       & Fibromyalgia     & \cite{guymer2019pharmacological} \\
Amlodipine   & High blood pressure     & \cite{donato2019black} \\
\rowcolor{black!20} Atorvastatin & High blood pressure & \cite{bubnova2019effects} \\
\bottomrule
\end{tabular}
\label{table:table3}
\end{table}

\section{Conclusion}
We utilize the existing drug-disease-protein interactions in form of graph structures for finding likely drug-disease interactions. Nonetheless, when using complex drug-disease-protein graph structures for drug repositioning, our main goal is the discovery of the hidden and unknown relations between the components' interactions and finally using these unknown correlations to facilitate the complex and time-consuming process of drug discovery.

\bibliographystyle{IEEEtran}
\bibliography{refs}

\begin{thebibliography}{10}
\providecommand{\url}[1]{#1}
\csname url@samestyle\endcsname
\providecommand{\newblock}{\relax}
\providecommand{\bibinfo}[2]{#2}
\providecommand{\BIBentrySTDinterwordspacing}{\spaceskip=0pt\relax}
\providecommand{\BIBentryALTinterwordstretchfactor}{4}
\providecommand{\BIBentryALTinterwordspacing}{\spaceskip=\fontdimen2\font plus
\BIBentryALTinterwordstretchfactor\fontdimen3\font minus
  \fontdimen4\font\relax}
\providecommand{\BIBforeignlanguage}[2]{{%
\expandafter\ifx\csname l@#1\endcsname\relax
\typeout{** WARNING: IEEEtran.bst: No hyphenation pattern has been}%
\typeout{** loaded for the language `#1'. Using the pattern for}%
\typeout{** the default language instead.}%
\else
\language=\csname l@#1\endcsname
\fi
#2}}
\providecommand{\BIBdecl}{\relax}
\BIBdecl

\bibitem{fda}
G.~Jin and S.~Wong, ``Toward better drug repositioning: prioritizing and
  integrating existing methods into efficient pipelines,'' \emph{Drug Discovery
  Today}, vol.~19, no.~5, pp. 637--644, 2014.

\bibitem{viagra}
R.~Renaud and H.~Xuereb, ``Erectile-dysfunction therapies,'' \emph{Nature
  Reviews Drug Discovery}, vol.~1, no.~9, pp. 663--664, 2002.

\bibitem{moa}
F.~Iorio, R.~Bosotti, E.~Scacheri, V.~Belcastro, P.~Mithbaokar, R.~Ferriero,
  L.~Murino, R.~Tagliaferri, N.~Brunetti-Pierri, A.~Isacchi, and
  D.~di~Bernardo, ``Discovery of drug mode of action and drug repositioning
  from transcriptional responses,'' \emph{Proc. of the National Academy of
  Sciences}, vol. 107, no.~33, pp. 14\,621--14\,626, 2010.

\bibitem{ong}
L.~Ong, B.~Cheung, Y.~Man, P.~Lau, and S.~Lam, ``Prevalence, awareness,
  treatment, and control of hypertension among united states adults
  1999–2004,'' \emph{Hypertension}, vol.~1, no.~49, pp. 69--75, 2007.

\bibitem{polypharmacy}
D.~Car, \emph{Polypharmacology in Drug Discovery}.\hskip 1em plus 0.5em minus
  0.4em\relax Wiley, 2012.

\bibitem{poly1}
W.~Zhang, Y.~Bai, Y.~Wang, and W.~Xiao, ``Polypharmacology in drug discovery: A
  review from systems pharmacology perspective,'' \emph{Curr Pharm Des.},
  vol.~22, no.~21, pp. 3171--3181, 2016.

\bibitem{skipgram}
T.~Mikolov, I.~Sutskever, K.~Chen, G.~Corrado, and J.~Dean, ``Distributed
  representations of words and phrases and their compositionality,'' in
  \emph{Proc. of {NIPS}}, 2013, pp. 3111--3119.

\bibitem{node2vec}
A.~Grover and J.~Leskovec, ``node2vec: Scalable feature learning for
  networks,'' in \emph{Proc. of {KDD}}, 2016, pp. 855--864.

\bibitem{deepwalk}
B.~Perozzi, R.~Al{-}Rfou, and S.~Skiena, ``Deepwalk: Online learning of social
  representations,'' in \emph{Proc. of {KDD}}, 2014, pp. 701--710.

\bibitem{nbne}
T.~Pimentel, A.~Veloso, and N.~Ziviani, ``Fast node embeddings: Learning
  ego-centric representations,'' in \emph{Proc. of {ICLR}}, 2018.

\bibitem{ehrt2016impact}
C.~Ehrt, T.~Brinkjost, and O.~Koch, ``Impact of binding site comparisons on
  medicinal chemistry and rational molecular design,'' \emph{Journal of
  Medicinal Chemistry}, vol.~59, no.~9, pp. 4121--4151, 2016.

\bibitem{lamb2006connectivity}
J.~Lamb, E.~D. Crawford, D.~Peck, J.~W. Modell, I.~C. Blat, M.~J. Wrobel,
  J.~Lerner, J.-P. Brunet, A.~Subramanian, K.~N. Ross \emph{et~al.}, ``The
  connectivity map: using gene-expression signatures to connect small
  molecules, genes, and disease,'' \emph{Science}, vol. 313, no. 5795, pp.
  1929--1935, 2006.

\bibitem{liu2010pharmmapper}
A.~Liu, S.~Ouyang, B.~Yu, Y.~Liu, K.~Huang, J.~Gong, S.~Zheng, Z.~Li, H.~Li,
  and H.~Jiang, ``Pharmmapper server: a web server for potential drug target
  identification using pharmacophore mapping approach,'' \emph{Nucleic acids
  research}, vol.~38, no.~2, pp. 609--614, 2010.

\bibitem{kumar2019drug}
A.~Kumar, K.~Ramaraju, R.~Singh, B.~Mittal, R.~Bhargava, and M.~Mittal, ``Drug
  delivery device for pharmaceutical compositions,'' Mar.~26 2019, uS Patent
  App. 10/238,803.

\bibitem{donner2018drug}
Y.~Donner, S.~Kazmierczak, and K.~Fortney, ``Drug repurposing using deep
  embeddings of gene expression profiles,'' \emph{Molecular Pharmaceutics},
  vol.~15, no.~10, pp. 4314--4325, 2018.

\bibitem{hu2009human}
G.~Hu and P.~Agarwal, ``Human disease-drug network based on genomic expression
  profiles,'' \emph{PloS one}, vol.~4, no.~8, p. e6536, 2009.

\bibitem{angles2008survey}
R.~Angles and C.~Gutierrez, ``Survey of graph database models,'' \emph{ACM
  Computing Surveys}, vol.~40, no.~1, p.~1, 2008.

\bibitem{belkin2002laplacian}
M.~Belkin and P.~Niyogi, ``Laplacian eigenmaps and spectral techniques for
  embedding and clustering,'' in \emph{Proc. of {NIPS}}, 2002, pp. 585--591.

\bibitem{kipf2016semi}
T.~N. Kipf and M.~Welling, ``Semi-supervised classification with graph
  convolutional networks,'' \emph{arXiv preprint arXiv:1609.02907}, 2016.

\bibitem{zitnik2018modeling}
M.~Zitnik, M.~Agrawal, and J.~Leskovec, ``Modeling polypharmacy side effects
  with graph convolutional networks,'' \emph{Bioinformatics}, vol.~34, no.~13,
  pp. i457--i466, 2018.

\bibitem{deepika2018meta}
S.~Deepika and T.~Geetha, ``A meta-learning framework using representation
  learning to predict drug-drug interaction,'' \emph{Journal of Biomedical
  Informatics}, vol.~84, pp. 136--147, 2018.

\bibitem{gao2018interpretable}
K.~Y. Gao, A.~Fokoue, H.~Luo, A.~Iyengar, S.~Dey, and P.~Zhang, ``Interpretable
  drug target prediction using deep neural representation.'' in \emph{Proc. of
  {IJCAI}}, 2018, pp. 3371--3377.

\bibitem{yamanishi}
Y.~Yamanishi, M.~Araki, A.~Gutteridge, W.~Honda, and M.~Kanehisa, ``Prediction
  of drug-target interaction networks from the integration of chemical and
  genomic spaces,'' \emph{Bioinformatics}, vol.~24, pp. 232--240, 2008.

\bibitem{zheng}
X.~Zheng, H.~Ding, H.~Mamitsuka, and S.~Zhu, ``Collaborative matrix
  factorization with multiple similarities for predicting drug-target
  interactions,'' in \emph{Proc. of {KDD}}, 2013, pp. 1025--1033.

\bibitem{xia}
Z.~Xia, L.~Wu, X.~Zhou, and S.~Wong, ``Semi-supervised drug-protein interaction
  prediction from heterogeneous biological spaces,'' \emph{BMC Systems
  Biology}, vol.~4, no.~6, 2010.

\bibitem{fingerprint}
D.~Duvenaud, D.~Maclaurin, J.~Aguilera{-}Iparraguirre,
  R.~G{\'{o}}mez{-}Bombarelli, T.~Hirzel, A.~Aspuru{-}Guzik, and R.~Adams,
  ``Convolutional networks on graphs for learning molecular fingerprints,'' in
  \emph{Proc. of {NIPS}}, 2015, pp. 2224--2232.

\bibitem{gcn}
W.~Hamilton, Z.~Ying, and J.~Leskovec, ``Inductive representation learning on
  large graphs,'' in \emph{Proc. of {NIPS}}, 2017, pp. 1024--1034.

\bibitem{chemistry}
J.~Gilmer, S.~Schoenholz, P.~Riley, O.~Vinyals, and G.~Dahl, ``Neural message
  passing for quantum chemistry,'' in \emph{Proc. of {ICML}}, 2017, pp.
  1263--1272.

\bibitem{menche}
J.~Menche, A.~Sharma, M.~Kitsak, S.~Ghiassian, M.~Vidal, J.~Loscalzo, and A.-L.
  Barabási, ``Uncovering disease-disease relationships through the incomplete
  interactome,'' \emph{Science}, vol. 347, p. 1257601, 2015.

\bibitem{biogrid}
A.~Chatr-aryamontri, B.-J. Breitkreutz, R.~Oughtred, L.~Boucher, S.~Heinicke,
  D.~Chen, C.~Stark, A.~Breitkreutz, N.~Kolas, L.~O'Donnell, T.~Reguly,
  J.~Nixon, L.~Ramage, A.~Winter, A.~Sellam, C.~Chang, J.~Hirschman,
  C.~Theesfeld, J.~Rust, M.~Livstone, K.~Dolinski, and M.~Tyers, ``The biogrid
  interaction database: 2015 update,'' \emph{Nucleic Acids Res.}, vol.~43, pp.
  D470--D478, 2015.

\bibitem{damian}
D.~Szklarczyk, J.~Morris, H.~Cook, M.~Kuhn, S.~Wyder, M.~Simonovic, A.~Santos,
  N.~Doncheva, A.~Roth, P.~Bork, L.~Jensen, and C.~von Mering, ``The string
  database in 2017: quality-controlled protein–protein association networks,
  made broadly accessible,'' \emph{Nucleic Acids Res.}, vol.~45, pp.
  D362--D368, 2017.

\bibitem{drugbank}
D.~Wishart, C.~Knox, A.~Guo, D.~Cheng, S.~Shrivastava, D.~Tzur, B.~Gautam, and
  M.~Hassanali, ``Drugbank: a knowledgebase for drugs, drug actions and drug
  targets,'' \emph{Nucleic Acids Res.}, vol.~36, pp. D901--D906, 2008.

\bibitem{ref5}
T.~Pimentel, A.~Veloso, and N.~Ziviani, ``Unsupervised and scalable algorithm
  for learning node representations,'' in \emph{Proc. of {ICLR}}, 2017.

\bibitem{ref8}
T.~Pimentel, R.~Castro, A.~Veloso, and N.~Ziviani, ``Efficient estimation of
  node representations in large graphs using linear contexts,'' in \emph{Proc.
  of {IJCNN}}, 2019, pp. 1--8.

\bibitem{ref6}
A.~Marczewski, A.~Veloso, and N.~Ziviani, ``Learning transferable features for
  speech emotion recognition,'' in \emph{Proc. {ACM} Multimedia}, 2017, pp.
  529--536.

\bibitem{ref3}
R.~de~Oliveira~Jr., A.~Veloso, A.~Pereira, W.~{Meira Jr.}, R.~Ferreira, and
  S.~Parthasarathy, ``Economically-efficient sentiment stream analysis,'' in
  \emph{Proc. of {ACM} {SIGIR}}, 2014, pp. 637--646.

\bibitem{tsne}
L.~v.~d. Maaten and G.~Hinton, ``Visualizing data using t-sne,'' \emph{Journal
  of machine learning research}, vol.~9, no. Nov, pp. 2579--2605, 2008.

\bibitem{ref10}
J.~Reis, A.~Correia, F.~Murai, A.~Veloso, and F.~Benevenuto, ``Explainable
  machine learning for fake news detection,'' in \emph{Proc. of {ACM}
  {WebSci}}, 2019, pp. 17--26.

\bibitem{shap}
S.~M. Lundberg and S.-I. Lee, ``A unified approach to interpreting model
  predictions,'' in \emph{Proc. of {NIPS}}, 2017, pp. 4765--4774.

\bibitem{hung2019migraine}
C.-I. Hung, C.-Y. Liu, C.-H. Yang, and S.-J. Wang, ``Migraine with active
  headache was associated with other painful physical symptoms at two-year
  follow-up among patients with major depressive disorder,'' \emph{PloS one},
  vol.~14, no.~4, p. e0216108, 2019.

\bibitem{enriquez2017cerebral}
A.~Enriquez-Marulanda, J.~Valderrama-Chaparro, L.~Parrado, J.~D. V{\'e}lez,
  A.~M. Granados, J.~L. Orozco, and J.~Qui{\~n}ones, ``Cerebral toxoplasmosis
  in an ms patient receiving fingolimod,'' \emph{Multiple sclerosis and related
  disorders}, vol.~18, pp. 106--108, 2017.

\bibitem{siciliano2019assessing}
M.~Siciliano, L.~Trojano, F.~Trojsi, M.~Monsurr{\`o}, G.~Tedeschi, and
  G.~Santangelo, ``Assessing anxiety and its correlates in amyotrophic lateral
  sclerosis: The state-trait anxiety inventory,'' \emph{Muscle \& nerve}, 2019.

\bibitem{fullerton2010rise}
C.~A. Fullerton, A.~B. Busch, and R.~G. Frank, ``The rise and fall of
  gabapentin for bipolar disorder: a case study on off-label pharmaceutical
  diffusion,'' \emph{Medical care}, vol.~48, no.~4, p. 372, 2010.

\bibitem{khalighi2016low}
H.~R. Khalighi, H.~Mortazavi, S.~M. Mojahedi, S.~Azari-Marhabi, and F.~M.
  Abbasabadi, ``Low level laser therapy versus pharmacotherapy in improving
  myofascial pain disorder syndrome,'' \emph{Journal of lasers in medical
  sciences}, vol.~7, no.~1, p.~45, 2016.

\bibitem{bosco2019clinical}
G.~Bosco, E.~Ostardo, A.~Rizzato, G.~Garetto, M.~Paganini, G.~Melloni,
  G.~Giron, L.~Pietrosanti, I.~Martinelli, and E.~Camporesi, ``Clinical and
  morphological effects of hyperbaric oxygen therapy in patients with
  interstitial cystitis associated with fibromyalgia,'' \emph{BMC urology},
  vol.~19, no.~1, p. 108, 2019.

\bibitem{guymer2019pharmacological}
E.~K. Guymer and G.~O. Littlejohn, ``Pharmacological treatment options for
  fibromyalgia,'' \emph{Prevention}, vol.~10, p.~00, 2019.

\bibitem{bubnova2019effects}
M.~Bubnova, D.~Aronov, and A.~Persiyanova-Dubrova, ``Effects of rosuvastatin
  and atorvastatin on blood pressure, cerebral blood flow, endothelial
  function, angiotensin ii in patients with ischemic stroke-complicated
  hypertension,'' \emph{Journal of Hypertension}, vol.~37, 2019.

\bibitem{donato2019black}
A.~Donato and K.~Brown, ``In black africans with hypertension, amlodipine-based
  therapy vs perindopril--hydrochlorothiazide improved bp control,''
  \emph{Annals of Internal Med.}, vol. 171, no.~2, pp. JC5--JC5, 2019.

\end{thebibliography}

\end{document}